\documentclass[11pt]{article}

\usepackage[preprint]{acl}

\usepackage{times}
\usepackage{latexsym}

\usepackage[T1]{fontenc}

\usepackage[utf8]{inputenc}

\usepackage{microtype}
\usepackage{inconsolata}
\usepackage{graphicx}

\usepackage[ruled,linesnumbered]{algorithm2e}

\usepackage{enumitem}

\title{ROC Analysis for Evaluating Translation Quality Estimation Systems}

\author{
\textbf{Evelyn Y. Garland\textsuperscript{1}} and
\textbf{Carola F. Berger\textsuperscript{2}}
\\
\textsuperscript{1}Acta Language Services, LLC,
\textsuperscript{2}CFB Scientific Translations LLC
\\
\small{
 \textbf{Correspondence:} \href{mailto:egarland@actalanguage.com}{egarland@actalanguage.com}
  }
}

\begin{document}
\maketitle

\begin{abstract}
The increasing use of automated translation quality estimation (QE) systems calls for practical, decision-oriented methods for evaluating their performance. We propose that Receiver Operating Characteristic (ROC) analysis is a useful approach for this purpose. Our study shows that ROC analysis not only produces results consistent with currently prevalent methods, but also offers several important advantages, including actionable performance insights that support business decision-making.
\end{abstract}

\section{Introduction}

As automated translation becomes more widely used, the number of systems designed to assess translation quality without relying on a reference translation has also grown significantly. These reference-free systems have been referred to by different names. Here, we refer to them as quality estimation (QE) systems. 

QE systems are used to support decisions involving large volumes of data, such as triaging for
translation workflow, preprocessing of data for model training, and prompt engineering and model evaluation. These and other high-impact use cases then pose an important question: How suitable is a specific QE system for the respective use case? In other words, how can the performance of the QE systems themselves be assessed?  

Studies evaluating QE system performance have been conducted within the framework of the WMT Machine Translation Metrics  and Quality Estimation shared tasks, most recently in~\cite{lavie-etal-2025-findings, freitag-etal-2024-llms, zerva-etal-2024-findings, freitag-EtAl:2023:WMT, blain-etal-2023-findings}. However, Receiver Operating Characteristic (ROC) analysis has not found widespread  use in WMT studies, which have generally emphasized other metrics.

Outside the WMT conference series, ROC curves have been used or considered when QE is framed as a binary detection, classification, or triage problem~\cite{lo-simard-2019-fully, blatz-etal-2004-confidence}, to a limited extent. 

A QE system may output quality scores, locations of errors (error span), estimates of post-editing effort, error-type classification, error severity ratings, or a combination of these. Many  QE systems are capable of producing at least quality scores that indicate the relative ranking of translations in terms of translation quality, and in this paper, we focus on these types of systems. 

This article attempts to provide general guidance on how ROC curves can be used to evaluate the performance of QE systems and support business-relevant decisions with the goal of promoting better evaluation practices.

\section{Receiver Operating Characteristic Analysis}

ROC analysis was developed in the 1940s by electrical engineers, but has since found widespread applications in many fields, including clinical research and machine learning. However, the full ROC analysis methodology appears not to have been widely applied to evaluate the performance of translation QE systems. Before explaining how to construct ROC curves and tables, we will review a few definitions to clarify the notation used in this paper. We assume that the text is divided into segments, as is standard in translation and natural language processing.

\subsection{Definitions and Review of Related Translation QE Metrics}\label{sec:definitions}
 
In natural language processing, QE systems can be used, among other things, to classify translation segments into two categories: error-free and error-containing segments. In this paper, we are concerned with locating segments that contain errors and therefore define the following:
\begin{eqnarray*}
Error  & = &  positive \, , \\
No\, \,error & = & negative \, .
\end{eqnarray*}

Let $P$ denote the number of segments containing errors (positives) and $N$ the number of error-free segments (negatives). Thus the {\em total number of segments} is $P + N$. This categorization forms the ground truth. Here, we do not distinguish whether an error-containing segment contains one error or many errors; it is still classified as {\em positive}, that is, containing at least one error.

If we then run a classifier (a QE system) over the translation segments, the classifier can be used to predict whether each segment contains an error or not. In Sections~\ref{sec:ROCsort} and ~\ref{sec:MQM} we show how QE system scores can be used for this classification.  Therefore, we have four possible outcomes: 
\begin{itemize}[nosep]
\item {\em TP, true positive}: an error-containing segment correctly classified as such,
\item  {\em FN, false negative}: an error-containing segment incorrectly classified as error-free,
\item {\em TN, true negative}: an error-free segment correctly classified as such,
\item {\em FP, false positive}: an error-free segment incorrectly classified as containing an error.
\end{itemize}
From these, we can construct a two-by-two {\em confusion matrix}, as is well known in the literature.

From the confusion matrix, we can calculate two metrics that are defined as follows:
\begin{eqnarray}
TPR & = & \frac{TP}{P} =  \frac{TP}{TP + FN} \, \label{TPReq} ,\\
FPR & = & \frac{FP}{N} =  \frac{FP}{FP + TN} \, \label{FPReq} .
\end{eqnarray}
That is, the {\em true positive rate, TPR}, is the ratio of true positives to the total number of positives, or, in our case, the number of error-containing segments that the QE system correctly identified as such divided by the total number of error-containing segments. Likewise, the {\em false positive rate, FPR}, is the number of false positives divided by the total number of negatives, or the ratio of error-free segments wrongly identified as containing at least one error to the total number of error-free segments.

Other commonly used metrics include:
\begin{eqnarray}
precision & = & \frac{TP}{TP + FP} \, , \\
recall & = & \frac{TP}{TP + FN} = TPR \,   .
\end{eqnarray}
In other words, the {\em true positive rate} is also called {\em recall}, because it is a measure of how completely the classifier "recalls" or identifies all positives. Another term for {\em TPR} or {\em recall} is {\em sensitivity}. {\em Precision}, as the name suggests, measures how precisely the QE system is able to correctly classify error-containing segments. {\em FPR} is related to another measure, {\em specificity}, by {\em FPR = $1 -$ specificity}. 

\subsection{ROC Curves for QE System Scores}\label{sec:ROCsort}

Now that we have clarified our notation, let us briefly summarize ROC analysis. For an in-depth review of ROC analysis, we refer the reader to ~\cite{FAWCETT2006861}. ROC graphs are two-dimensional graphs with {\em TPR} plotted on the y-axis against {\em FPR} on the x-axis for varying threshold values of the classifier. Such a two-dimensional graph illustrates the relative trade-offs between benefits (true positives) and costs (false positives). In our case, the benefits are correctly classified error-containing segments, because these can then be corrected by a human or possibly an automated system or correctly excluded in a data preprocessing step for model training. False positives create additional costs, because these have to be processed further in a translation workflow even though they are, in fact, error-free, or unnecessarily excluded from a cleaned dataset, even though they are actually usable.

\begin{figure}[htb!]
\centering
 \includegraphics[width=0.85\columnwidth]{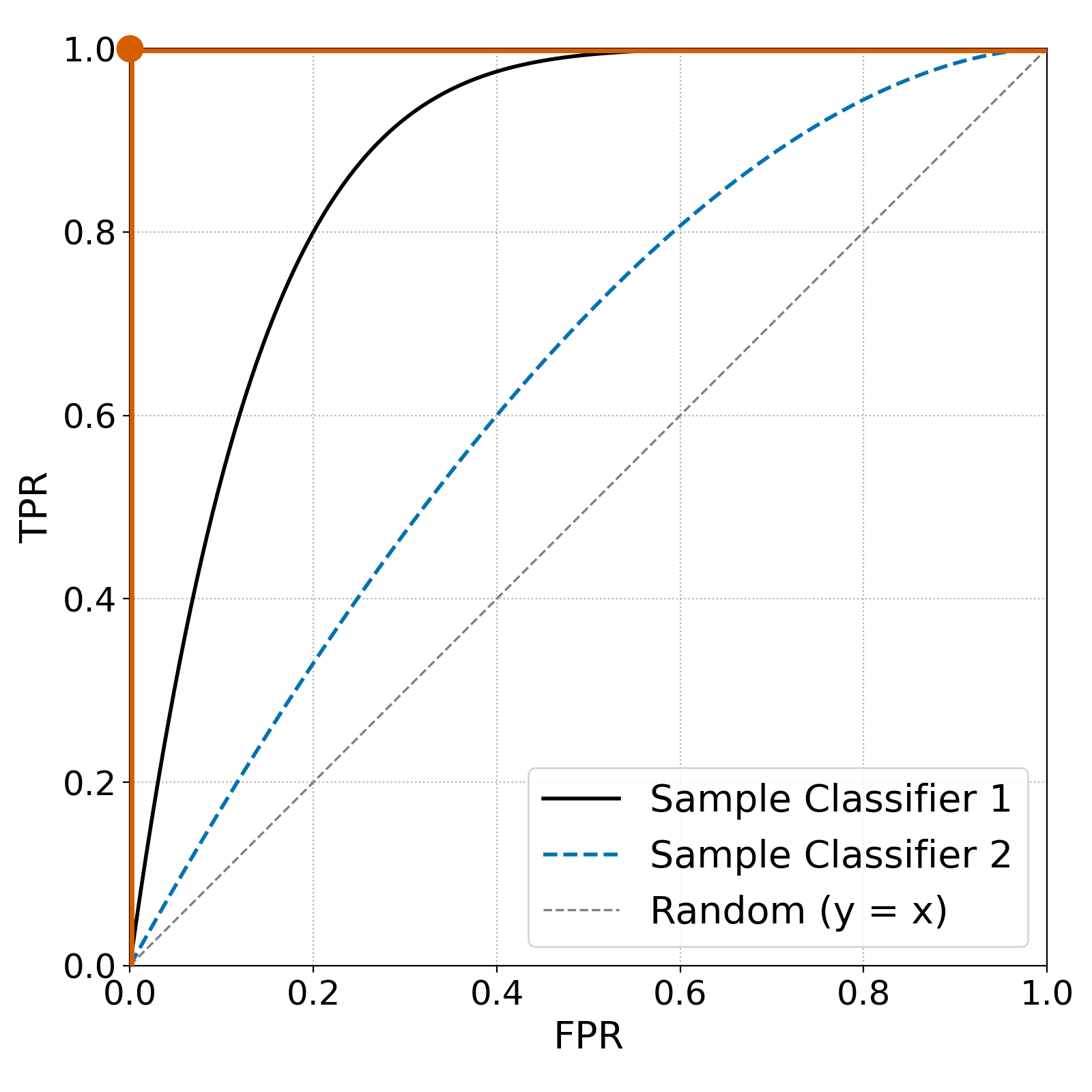} 
  \caption{Idealized ROC curves for two sample classifiers, where classifier 1 performs relatively better than classifier 2.}
  \label{fig:ROCsample}
\end{figure}

Figure~\ref{fig:ROCsample} shows idealized curves in ROC space for continuous classifiers. In reality, we have a finite set of data points or classified segments, and our ROC curves will be step functions or piecewise linear functions, depending on the data and classifier. In the figure, the dashed diagonal line, {\em TPR = FPR}, illustrates the behavior of a random classifier that guesses correctly half of the time and wrongly the other half of the time. The orange line that passes through point (0,1) in ROC space denotes a perfect classifier that gets it right every time. The closer a curve is to this ideal curve, the better the classifier. Thus, in Figure~\ref{fig:ROCsample}, the solid black curve is generally a better classifier than the dashed blue curve for this particular dataset.

Let us now explain how to obtain ROC curves for our case, where we have a sample dataset of translated segments, annotated or post-edited by humans in some fashion. In this paper, this forms our {\em gold standard}. Generally speaking, a gold standard is the benchmark that is the best available under reasonable conditions. With the help of this gold standard, we can categorize the segments into two classes: error-free (negatives) or containing one or more errors (positives). This categorization into positives and negatives forms the {\em ground truth}. In Section~\ref{sec:MQM}, we show how to obtain this ground truth from MQM and MQM-related scores. 

We also have a score for each segment produced by a QE system.  Usually, these scores tend to be either discrete or near-continuous, say, an integer between 0 and 100, or decimal numbers, say, between 0.0 and 1.0, with fixed decimal precision, but they can also be unbounded. These scores form the classifier threshold values that we sweep to produce the ROC curve. Standard ROC applications and software implementations assume that {\bf a higher classifier score} indicates a {\bf more positive instance}. However, some QE systems output a higher score for segments that are more "negative" (more "error-free," i.e., better translation quality), the exact reverse of the expected input into the algorithm. For these scores, the standard algorithm still applies if the scores are negated. Following this standard algorithm, detailed, for example, in Algorithm 1 of~\cite{FAWCETT2006861}, we obtain an ROC curve, which is usually a two-dimensional step function.

For two or more data points with the same QE score, the ROC curve will not be a step function, but rather a piecewise linear curve, because we adopt the {\em expected} performance approach of~\cite{FAWCETT2006861}. In principle, differently translated segments should produce different QE system scores. However, with discrete classifiers or due to rounding of floating point numbers, it can happen that different segments result in the same classifier score, thus producing two or more different data points for the same score threshold. For these tied thresholds, we add all true and false positives to produce one {\em TPR} and {\em FPR} value, respectively, for each unique QE score threshold. This then modifies the step function into a piecewise linear curve. We will see examples in our results below.

\subsection{Area Under the Curve}

Another metric that can be computed from ROC curves is the {\em area under the curve (AUC}, elsewhere also referred to as {\em AUROC}), which condenses the two-dimensional information contained in ROC curves into a single value. It measures the classifier's ability to distinguish between positives and negatives (errors and no errors) for all possible thresholds. AUC ranges from 0 to 1, where 0.5 is no better than random guessing. We refer to~\cite{FAWCETT2006861} for instructions on how to compute the AUC value via the trapezoidal area under the curve. 

Note that different ROC curves with different shapes can have the same AUC value if they intersect each other. The reason for this is that the AUC value condenses the two-dimensional ROC information into a single, lossy value.

\subsection{Bootstrap Resampling to Assess Statistical Significance}\label{sec:bootstrap}

Determining statistical significance is useful for assessing the uncertainty in the performance of a single QE system and in the comparison of two or more QE systems. One way to visualize this uncertainty is through the confidence band of an ROC curve.  To assess the confidence band of the ROC curve, we use {\em nonparametric stratified bootstrap resampling} (see, e.g.,~\cite{Efron:1979bxm}), a standard algorithm implemented in many statistical software programs. We select this method because the underlying distributions are unknown, in particular for LLM-based QE systems evaluating LLM-based machine translation systems. Confidence intervals for AUC values are obtained analogously. The algorithm is explained in more detail in Appendix~\ref{app:bootstrap}.

It should be noted that these confidence bands and intervals do not take into account various biases, for example, variability in the ground truth due to inter-annotator disagreement. 

\section{Study Methodology}

\subsection{Data Selection}\label{sec:dataselection}

We wanted to study public datasets with established gold standard ratings and published analyses to verify against. Therefore, we selected data from past Conferences on Machine Translation (WMT) that were available in a public GitHub repository~{\tt https://github.com/google\-research/mt-metrics-eval}, licensed under Apache v2.0. Specifically, we chose data from WMT23~\cite{kocmi-EtAl:2023:WMT}, the Eighth Conference on Machine Translation, because this was the most recent complete dataset with full MQM annotation available.\footnote{For proprietary reasons, the WMT24 data in the repository are incomplete compared to the published articles. The WMT25 data did not contain a sufficient number of annotated segments per language pair for our analysis.} 

We selected the data from the MT Metrics Evaluation subtask~\cite{freitag-EtAl:2023:WMT}, which also included a QE system component, for the language pairs Chinese into English and English into German. For Chinese into English, there were 1,177 total segments with human gold standard annotation available, and for English into German a total of 460 (cf. Table 3 of~\cite{freitag-EtAl:2023:WMT}). We selected three different machine translation (MT) systems and four different QE systems for each language pair. 

We wanted to study whether the ROC analysis could reproduce the published WMT23 Metrics Shared Task rankings for reference-free systems on a variety of MT systems. Therefore, we selected the systems listed first and last for our two language directions in Table 6 of~\cite{kocmi-EtAl:2023:WMT}, as well as one listed in the middle. For Chinese into English, the three selected systems were Lan-BridgeMT, ONLINE-A, and ANVITA. For English into German, we chose GPT4-5shot, Lan-BridgeMT, and AIRC. For details on these systems, we refer to~\cite{kocmi-EtAl:2023:WMT}.

For the selection of QE systems, we wanted to choose two systems with better relative performance and two with worse relative performance, as determined by the WMT23 Metrics Shared Task.\footnote{Unfortunately, data for the WMT23 Quality Estimation Shared Task~\cite{blain-etal-2023-findings} were not available in the referenced GitHub repository.} Additionally, we wanted to study at least one LLM-based system among these as well as one of the systems the WMT group chose as baselines, and all selected systems had to be reference-free to qualify as QE systems. We selected the QE systems based on the Pearson rankings in Table 9 of~\cite{freitag-EtAl:2023:WMT} for Task 3 (en$\rightarrow$de) and Task 9 (zh$\rightarrow$en), respectively. For Chinese into English, our QE system choices were MetricX-23-QE, an encoder-decoder-based model that was not a Comet derivative, CometKiwi as the baseline system, and prismSrc and Random-sysname as representatives at the bottom of the WMT ranking. Similarly, for English into German, the QE systems chosen were GEMBA-MQM as an LLM-based system, CometKiwi as the baseline system, and again prismSrc and Random-sysname. Again, for details on these systems, we refer to~\cite{freitag-EtAl:2023:WMT}. 

\subsection{ROC Analysis}\label{sec:MQM}

For all of the aforementioned MT systems, a "merged" MQM-derived score produced by human annotators was available at the segment level, which we used as the gold standard. As described in~\cite{freitag-EtAl:2023:WMT}, the human annotators used a scheme based on MQM with major, minor, and neutral error severities. Major errors were weighted with 25 points for non-translation and 5 points otherwise; minor errors were weighted with 0.1 points for fluency or punctuation issues and 1 point in all other cases. Unlike the current standard MQM scoring model~\cite{lommel2024multirangetheorytranslationquality}, these points were not weighted by word count, but simply added per segment. In the GitHub repository (for round 1, see below), the scores were reported as negative values per segment  in the "merged" MQM ratings. Here, the term "merged" indicates that several different annotators rated different segments, and these ratings were combined into one single set of gold standard ratings. These ratings were repeated by other raters, for a total of three rounds with one rating per segment each. Below we use data from round 1 only, because the official WMT23 results are based on round 1 ratings. 

To categorize these ratings into the two classes, error-containing (positive) and error-free (negative), we used two different severity cutoffs, to see whether the inclusion of minor errors or the focus on major errors only would change our results. We used a severity cutoff of $<0$ error points to account for all errors in the classification, including the most minor ones. We also ran a separate analysis with a severity cutoff of $<= -5$, to allow several minor errors up to the cutoff, but no major errors, with the WMT23 error weighting as explained above. Based on these cutoffs, each segment was then categorized according to the human annotations with a ground-truth label of error-free (negative) or error-containing (positive).

We computed ROC curves and AUC values using the Python package {\tt scikit-learn}~\cite{scikit-learn}, supplemented by manual checks with Microsoft Excel. For QE systems where higher QE scores indicate better segments, that is, less "positive" segments (errors), we used the negatives of the QE scores as inputs for the ROC and AUC computation, because the package expects higher scores to indicate a higher likelihood that the instance is positive. We implemented the computation of the confidence bands with the bootstrap resampling method described in Section~\ref{sec:bootstrap} in Python. For the resampling algorithm, we used $B = 1,000$ iterations, because we found empirically that a larger number of iterations did not significantly alter the confidence bands and values.

\section{ROC Analysis for QE System Evaluation}

\subsection{Comparing ROC Analysis with Currently Prevalent Approaches}\label{sec:WMTrepro}

\begin{figure*}[bht!]
  \centering
  \begin{minipage}{0.47\textwidth}
    \centering
    \includegraphics[width=0.9\textwidth]{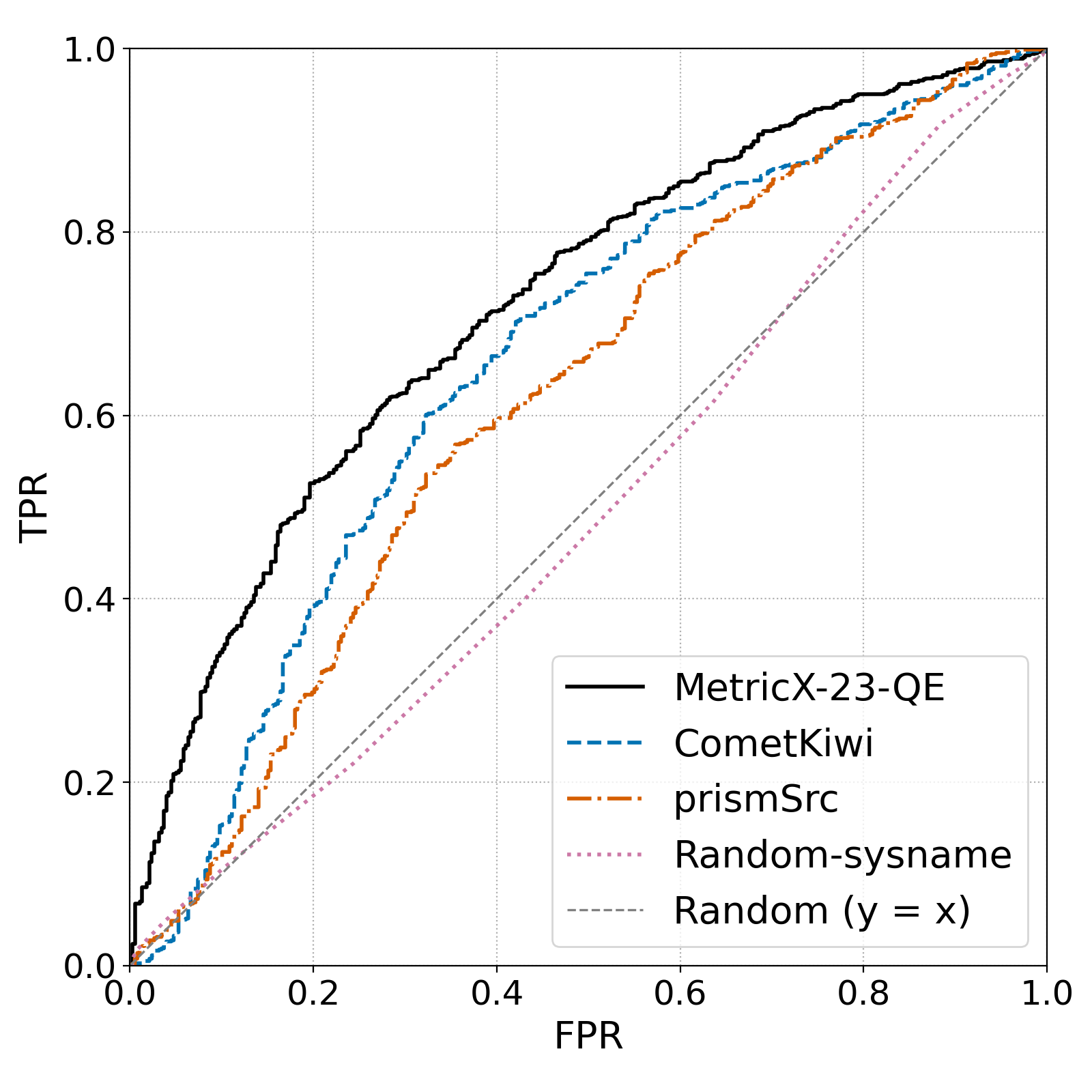}
    \caption{ROC curves for the WMT23 zh$\rightarrow$en dataset of 1,177 segments using the Lan-BridgeMT machine translation system classified with the strict severity cutoff $< 0$ to yield $P = 799$ error-containing segments.  }
     \label{fig:zhLanbridge}
  \end{minipage}
  \hfill
  \begin{minipage}{0.47\textwidth}
    \centering
    \includegraphics[width=0.9\textwidth]{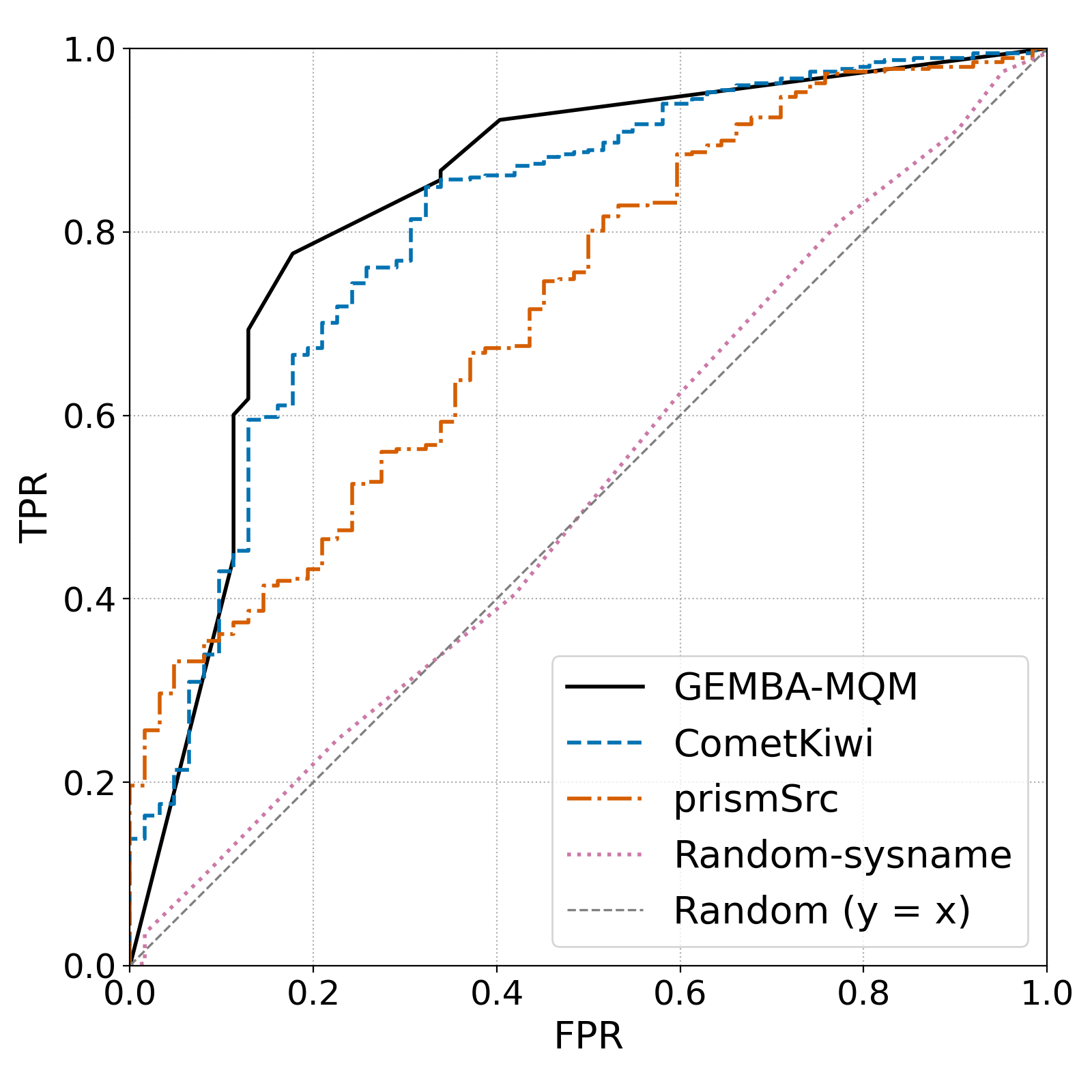}
    \caption{ROC curves for the WMT23 en$\rightarrow$de dataset of 460 segments using the AIRC machine translation system classified with the strict severity cutoff $< 0$ to yield $P = 398$ error-containing segments.}
    \label{fig:deAIRC}
  \end{minipage}
\end{figure*}

\begin{figure*}[htb!]
  \centering
  \begin{minipage}{0.47\textwidth}
    \centering
    \includegraphics[width=0.9\textwidth]{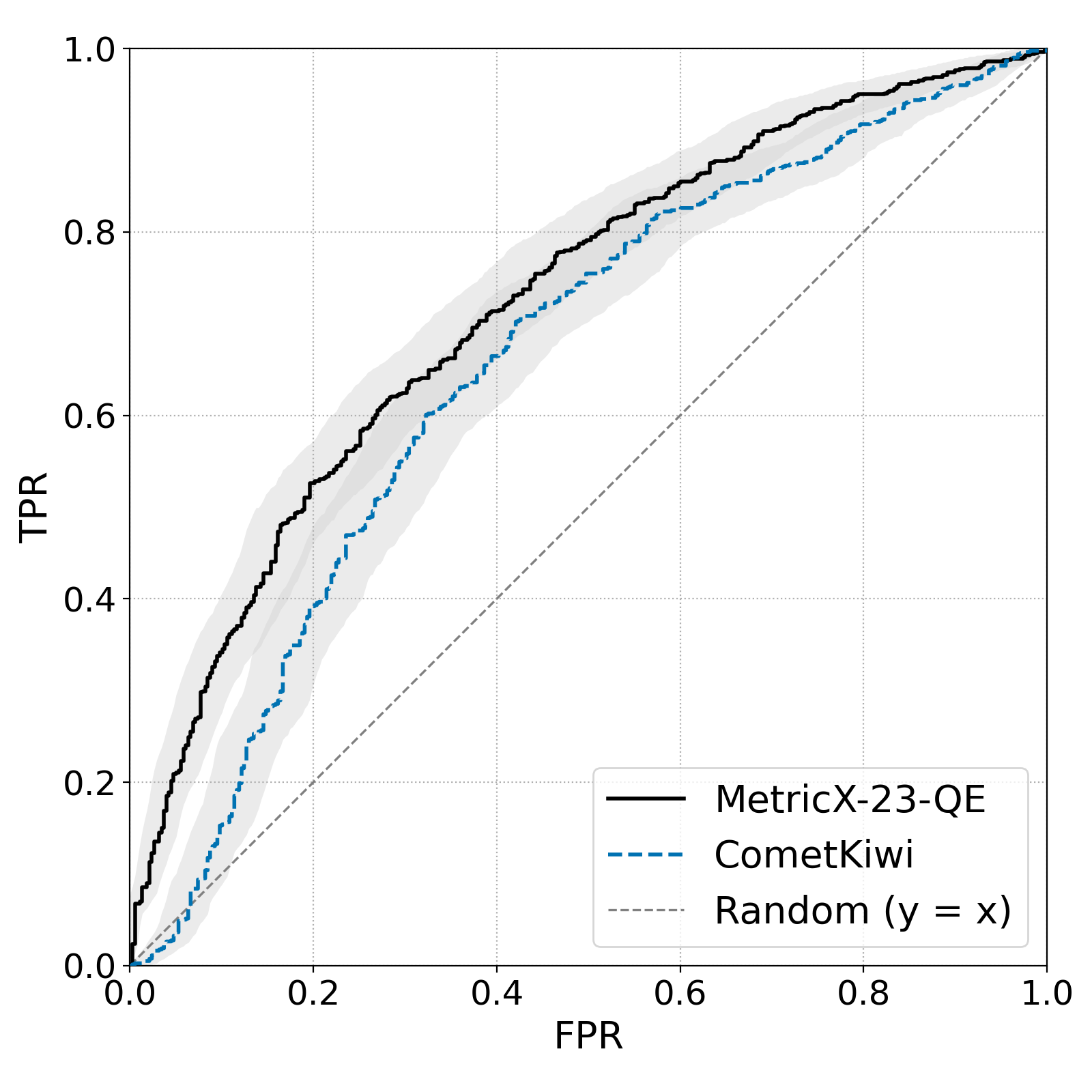}
    \caption{Same as Figure~\ref{fig:zhLanbridge} for zh$\rightarrow$en with confidence bands, but without the curves for prismSrc and Random-sysname. }
     \label{fig:zhLanbridgeconfidence}
  \end{minipage}
  \hfill
  \begin{minipage}{0.47\textwidth}
    \centering
    \includegraphics[width=0.9\textwidth]{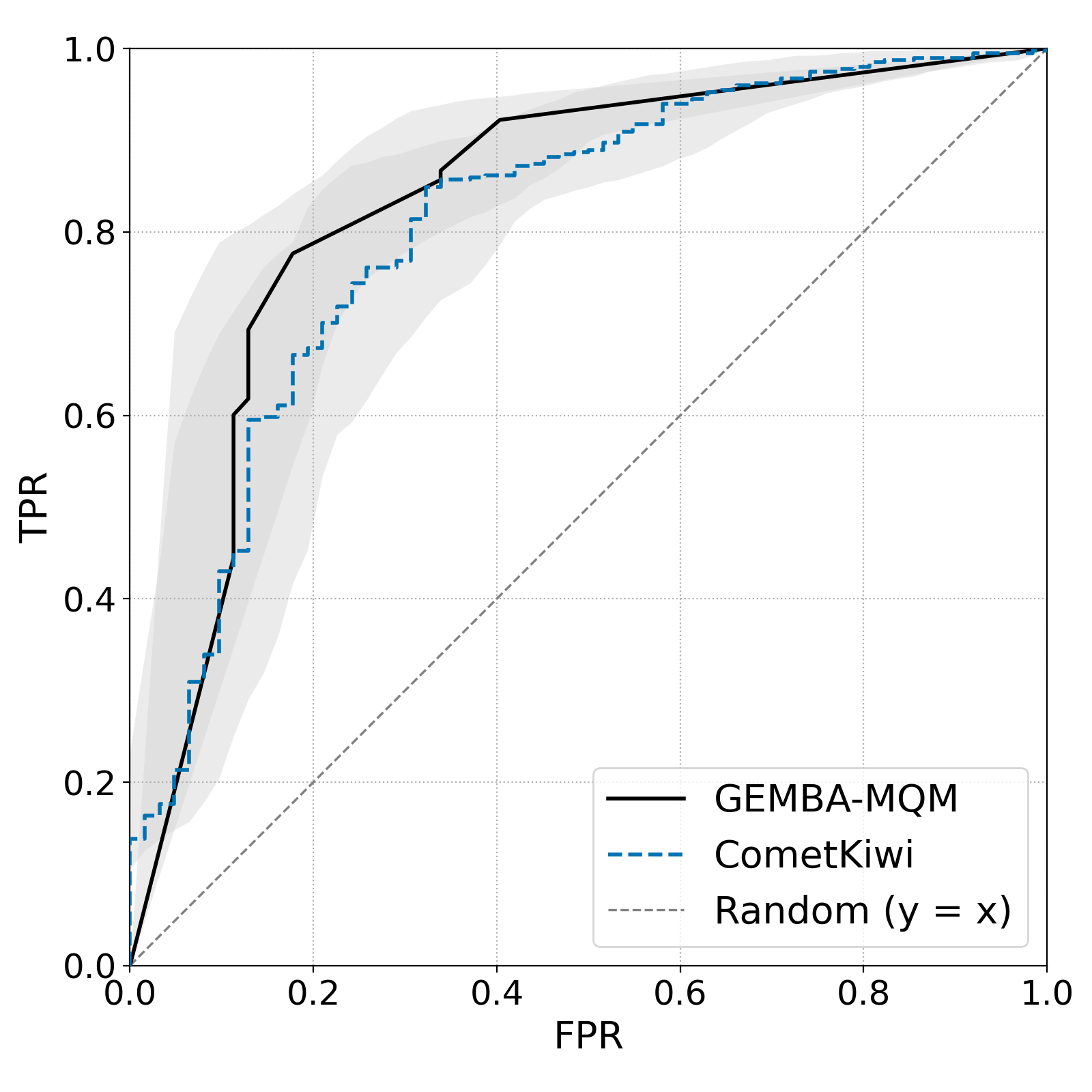}
    \caption{Same as Figure~\ref{fig:deAIRC} for en$\rightarrow$de with confidence bands, but without the curves for prismSrc and Random-sysname. }
    \label{fig:deAIRCconfidence}
  \end{minipage}
\end{figure*}

One reason to choose the public data available from WMT23 was to be able to compare our results with the findings published in~\cite{freitag-EtAl:2023:WMT}. Figures~\ref{fig:zhLanbridge} and~\ref{fig:deAIRC} show sample ROC curves for the Chinese into English and the English into German datasets, respectively, with the $<0$ severity cutoff. Figure~\ref{fig:zhLanbridge} shows the ROC curves for the four QE systems chosen in Section~\ref{sec:dataselection} that estimated the quality for the Lan-BridgeMT translation system for zh$\rightarrow$en. Similarly, Figure~\ref{fig:deAIRC} shows the ROC curves for the four QE systems chosen in Section~\ref{sec:dataselection} that estimated the quality for the AIRC translation system for en$\rightarrow$de. Table~\ref{table:AUCzh} lists all computed AUC values including their 95\% confidence intervals for all combinations of our chosen MT and QE systems for zh$\rightarrow$en, and Table~\ref{table:AUCde} lists the corresponding values for en$\rightarrow$de. To obtain the ROC graphs and AUC values, we followed the procedure detailed in Section~\ref{sec:MQM}.

\begin{table*}[tb!]
  \centering
  \begin{tabular}{l|lll}
    \hline
   & \textbf{Lan-BridgeMT} & \textbf{ONLINE-A} & \textbf{ANVITA} \\
    \hline
    \textbf{MetricX-23-QE} & $0.721 \pm 0.032$ & $0.812 \pm 0.030$ & $0.813 \pm 0.034$ \\
    \textbf{CometKiwi} & $0.657 \pm 0.033$ & $0.775 \pm 0.034$ & $0.794 \pm 0.034$ \\
    \textbf{prismSrc} & $0.618 \pm 0.035$ & $0.703 \pm 0.042$ & $0.744 \pm 0.044$ \\
    \textbf{Random-sysname} & $0.494 \pm 0.036$ & $0.528 \pm 0.041$ & $0.536 \pm 0.044$ 
       \\\hline
  \end{tabular}
\caption{AUC values and their 95\% confidence intervals for the various chosen MT and QE systems for zh$\rightarrow$en.}
  \label{table:AUCzh}
\end{table*}

\begin{table*}[bt!]
  \centering
  \begin{tabular}{l|lll}
    \hline
   & \textbf{GPT4-5shot} & \textbf{Lan-BridgeMT} & \textbf{AIRC} \\
    \hline
    \textbf{GEMBA-MQM} & $0.690 \pm 0.041$ & $0.801 \pm 0.046$ & $0.833 \pm 0.061$ \\
    \textbf{CometKiwi} & $0.715 \pm 0.050$ & $0.750 \pm 0.057$ & $0.808 \pm 0.064$ \\
    \textbf{prismSrc} & $0.641 \pm 0.055$ & $0.660 \pm 0.062$ & $0.719 \pm 0.064$ \\
    \textbf{Random-sysname} & $0.495 \pm 0.054$ & $0.526 \pm 0.064$ & $0.515 \pm 0.076$ 
       \\\hline
  \end{tabular}
\caption{AUC values and their 95\% confidence intervals for the various chosen MT and QE systems for en$\rightarrow$de.}
  \label{table:AUCde}
\end{table*}

Upon comparing with Table 9 in~\cite{freitag-EtAl:2023:WMT}, specifically Task 9 for zh$\rightarrow$en and Task 3 for en$\rightarrow$de, we observe that our ROC curves and corresponding AUC values closely match the results of WMT23 for different MT systems. We refrain from showing similar ROC curves for the other combinations of language pairs and MT systems, but they can be obtained from the authors upon request. Qualitatively, we are able to reproduce the relative performance results shown in Table 9 of~\cite{freitag-EtAl:2023:WMT} for the selected systems. 

 We repeated the computations with the less strict severity cutoff to allow a certain amount of minor errors (see Section~\ref{sec:MQM}). The results are very similar to those for the strict severity cutoff, and we refrain from showing them here. 

When looking at Tables~\ref{table:AUCzh} and~\ref{table:AUCde}, it should be noted that AUC values and ROC curves of one particular QE system evaluating several MT systems should {\em not} be used to rank the performance of the MT systems themselves. We therefore refrain from showing ROC curves of one QE system for several different MT systems in one graph, because this would be a misleading illustration. What indicates the performance of an MT system is the QE score, not the QE system's ability to separate the positives from the negatives for that MT system. Indeed, AUC values and ROC curves suggest which MT systems' outputs are more challenging for the QE system, not which MT systems produce better translations. The fact that the AUC values seem to decrease for better-performing MT systems (as ranked by WMT23) suggests that as MT systems continue to improve, they will likely be more challenging for QE systems, and QE systems therefore should continue to improve as well.

In terms of statistical significance, we show the same ROC curves for the same machine translation datasets for the two better-performing QE systems of Figures~\ref{fig:zhLanbridge} and~\ref{fig:deAIRC}, that is, MetricX-23-QE and CometKiwi for zh$\rightarrow$en as well as GEMBA-MQM and CometKiwi for en$\rightarrow$de, but now including confidence bands in Figures~\ref{fig:zhLanbridgeconfidence} and~\ref{fig:deAIRCconfidence}, respectively. We can easily see that the larger dataset for Chinese into English results in narrower confidence bands compared to English into German, as expected.

We can therefore conclude that ROC analysis is indeed a viable option to evaluate the relative performance of different QE systems when benchmarked against the same gold standard.

\subsection{Advantages of ROC Analysis}

ROC analysis is more than an alternative to currently prevalent methods for evaluating QE system performance. It provides richer information that helps users understand system behavior and make informed business decisions.

\subsubsection{Full Spectrum Performance}\label{sec:addtlinfo}

The ROC curve shows QE system performance over the entire spectrum of QE score thresholds that may be used to distinguish error-containing segments from error-free segments. This means that evaluation is not tied to a single, arbitrarily selected threshold. Instead, we can examine QE system performance across all thresholds and identify the threshold that is most appropriate for a given task. The selection of QE score thresholds is discussed further in Appendix~\ref{sec:mainthresholds}.

\subsubsection{Early Retrieval}

Because the ROC curve displays performance across the full QE score spectrum, or equivalently across the full range of false positive rates ({\em FPR}s), it also allows us to examine classifier behavior in different parts of that range. For example, when the task is to identify segments containing translation errors, it may be especially desirable for a QE system to retrieve such segments early, that is, in the low-{\em FPR} region where the QE scores are the worst. Performance in this region can be particularly informative for practical applications where false positives are costly or where only a very limited number of segments can be flagged for review. 

In other use cases, such as selecting high-quality translation segments from a large pool of data for model training purposes, the objective may instead be to identify error-free segments. In such cases, the portion of the ROC curve associated with better QE scores, or equivalently the high-{\em FPR} region, may be of greater interest, particularly when false negatives are costly or when only a small subset of the available data is needed.

\begin{figure}[htb!]
\centering
 \includegraphics[width=0.9\columnwidth]{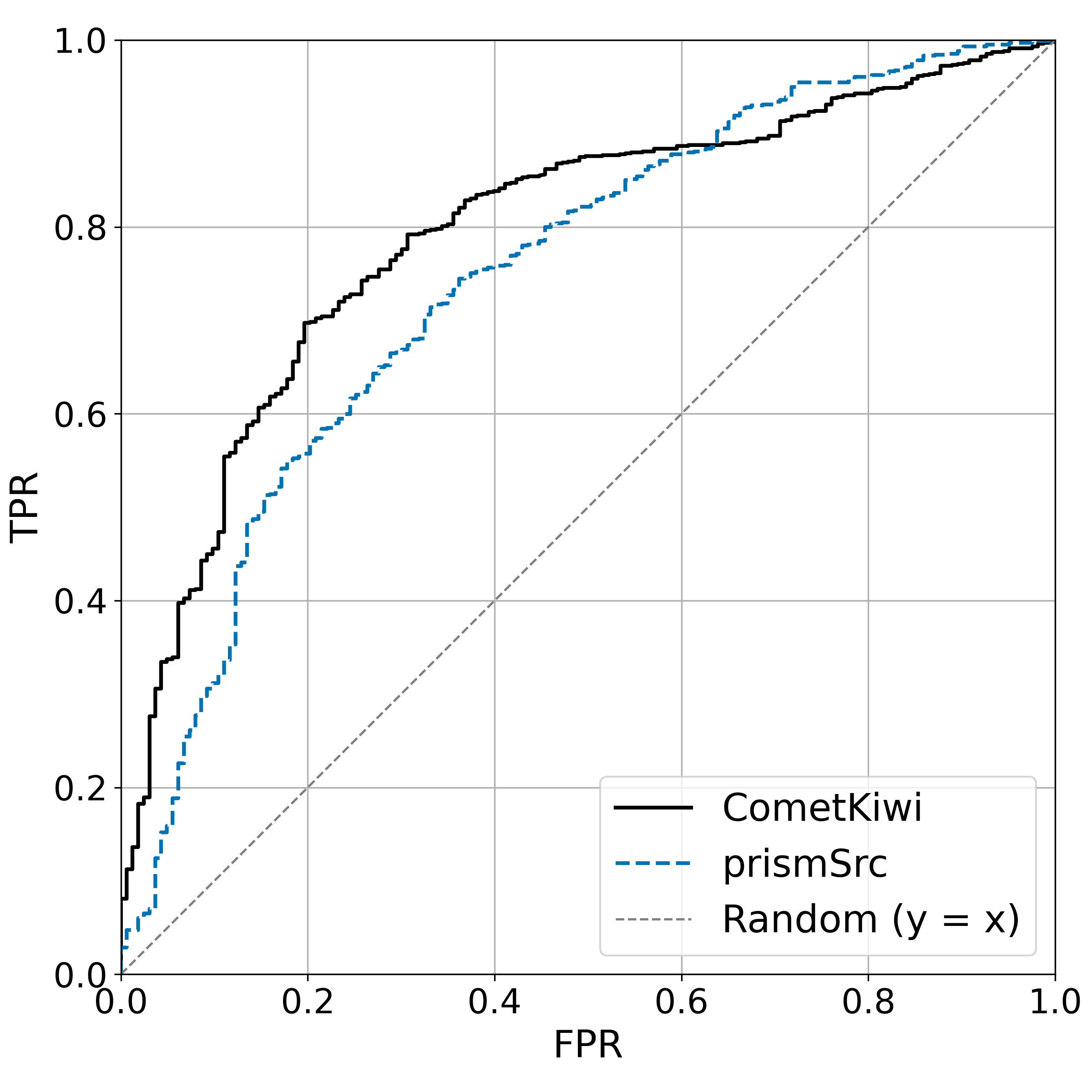} 
  \caption{Performance of two QE systems, CometKiwi and prismSrc, applied to the output of the same MT system, ANVITA, on the WMT23 zh$\rightarrow$en dataset.}
  \label{fig:twoQEsystems}
\end{figure}

Figure~\ref{fig:twoQEsystems} illustrates the performance of two QE systems applied to the same MT system. When the task is early retrieval of positive instances, that is, segments containing errors, CometKiwi performs better. By contrast, when the task is early retrieval of negative instances, that is, error-free segments, prismSrc is superior. If interest is restricted to a particular early-retrieval region, it is possible to compute and compare the AUC over that region rather than over the entire QE score range ~\cite{RICHARDSON2024100994}. If interest is in the region in the vicinity of where the two curves cross, we refer to the discussion on the ROC convex hull in~\cite{FAWCETT2006861}.

\subsubsection{Combining QE Systems}

ROC analysis also makes it possible to consider combinations of two or more QE systems in order to optimize performance. In the example above, where the ROC curves of two QE systems cross, the systems could be combined so that one is used in the low-{\em FPR} region and the other in the high-{\em FPR} region. Such a strategy could yield superior performance across the full QE score range, beyond what either system can achieve on its own.

\subsubsection{Visualizing {\em FNR-FPR} Trade-Off}

Another important advantage of ROC analysis is that it makes the trade-off between false negative rate ({\em FNR}) and false positive rate ({\em FPR}) explicit. Ideally, a QE system would classify all segments correctly, never misclassifying an error-containing segment as error-free or an error-free segment as error-containing. In reality, however, QE systems are imperfect and almost inevitably produce false negatives, false positives, or both. {\em FNR} and {\em FPR} depend on the QE score threshold: changing the threshold changes the balance of {\em FNR} and {\em FPR}. Although the general objective is to minimize both, a given QE system typically cannot reduce one without increasing the other. 

ROC analysis provides a clear framework for examining this relationship. Because the false negative rate ({\em FNR}) is related to the true positive rate ({\em TPR}) by
\begin{displaymath}
FNR = 1 - TPR \, ,
\end{displaymath}
movement along the ROC curve shows how gains in {\em TPR}, and thus reductions in {\em FNR}, are accompanied by increases in {\em FPR}, and vice versa. The ROC curve visualizes this relationship.

\begin{figure}[htb!]
\centering
  \includegraphics[width=\columnwidth]{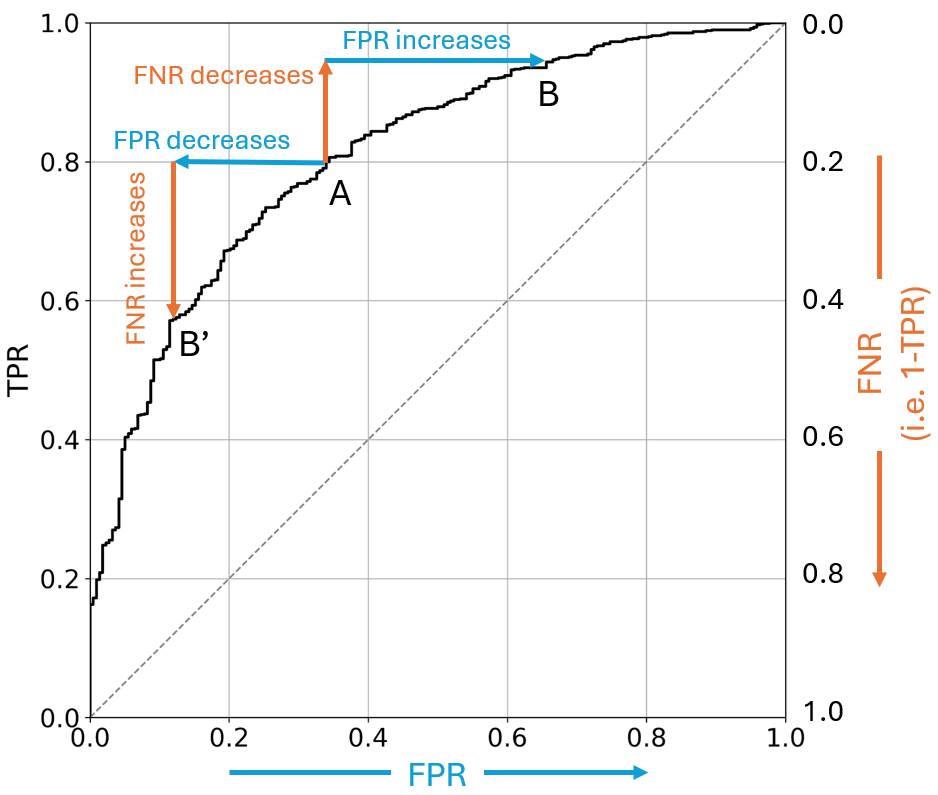} 
  \caption{This graph illustrates the trade-off between {\em FNR} and {\em FPR} on a sample ROC curve, with a second vertical axis added to the right indicating {\em FNR}. Moving along the ROC curve to the right (e.g., from point A to point B), {\em FNR} decreases and {\em FPR} increases. Conversely, moving to the left along the ROC curve (e.g., from point A to point B'), {\em FNR} increases and {\em FPR} decreases.}
  \label{fig:tradeoff}
\end{figure}

As illustrated in Figure~\ref{fig:tradeoff}, for a given QE system, if we start at point A and want to lower the {\em FNR}, we can move the operating point from A to B, but this will result in a higher {\em FPR}. Conversely, if we want to reduce the {\em FPR}, we can move from A to B', but at the cost of increasing the {\em FNR}. 

\subsubsection{Insights Into {\em FN-FP} Trade-Off}

The {\em FNR}-{\em FPR} relationship discussed above is particularly useful because the trade-off between these rates can be translated into absolute counts once the number of instances in each class ($P$ and $N$) is known. In ROC analysis, $P$ and $N$ are indeed known because they can be tallied from the ground truth. It then becomes possible to estimate how many additional false positives ({\em FP}s) would result from reducing false negatives ({\em FN}s) over a given portion of the ROC curve for a given QE system. That is to say, how many additional {\em FP}s, i.e., error-free segments that are "thought" to be error-containing by the QE system, would be incurred in order to reduce the number of {\em FN}s, i.e., error-containing segments missed by the QE system. Similarly, given the ROC curve and class ratio, we can calculate how many more {\em FN}s would result in order to reduce {\em FP}s to a certain level for a given QE system. This has direct implications for business decision-making and risk management, as discussed in greater detail in Appendix~\ref{sec:optimum}.

Note that {\em FPR} and {\em FNR} should not be confused with {\em FP} and {\em FN}, see Section~\ref{sec:definitions}.

\subsubsection{Minimal Score Manipulation When Comparing Systems}

Different QE systems often produce scores on different scales with unknown underlying distributions. 
An important property of an ROC graph is that it reflects a classifier's ability to assign discriminative {\em relative} scores to instances, i.e., segments in the case of QE. A classifier need not produce accurate, calibrated probability estimates; rather, it need only rank instances in a way that effectively separates positive from negative cases~\cite{FAWCETT2006861}. In other words, when using ROC analysis to evaluate QE system performance, we do not need the scores produced by different QE systems to be on the same scale, have the same probability distribution, or be indicative of the true probability that a segment contains errors. Neither normalization nor calibration is required for ROC analysis.

It should be noted that when the systems being compared use opposite scoring directions---for example, one where a higher score indicates better translation quality and another where a higher score indicates poorer translation quality---the scores must first be converted so that they change in the same direction. As explained in Section~\ref{sec:ROCsort}, this can be done simply by negating scores that change in one direction while leaving the others as is.

\section{Conclusions}

Given the increasing use of translation quality estimation (QE) systems across a wide range of applications, it is essential to evaluate the performance of QE systems themselves, both individually and in comparison to one another. In this paper, we have demonstrated how ROC analysis can be applied to the evaluation of translation QE systems. Using the ROC method, we were able not only to reproduce the relative ranking patterns observed in selected WMT23 results, but also to show that ROC analysis provides rich, actionable, and business-relevant insights beyond those offered by currently prevalent QE evaluation methods individually. 

\section{Limitations}

{\bf Mixed-domain data. }The WMT dataset includes source documents from multiple domains. Although each segment is labeled by domain, we used the full dataset without domain-based separation for two reasons. First, the official WMT23 Metrics Shared Task rankings of QE systems were based on all domains combined. Second, the number of segments in each individual domain was relatively small, with fewer than 400 segments in all but one domain.

Ideally, QE systems should be evaluated separately by domain, because text characteristics, error patterns, and the relationship between QE scores and actual translation quality may vary across domains. Domain-specific evaluation would therefore provide a more fine-grained view of system performance, but the available domain-level sample sizes were too limited for robust analysis in this study.

{\bf Gold standard based on a single set of ratings.} For many segments, WMT23 included three sets of ratings provided by different raters. However, the official WMT23 Metrics Shared Task rankings were based on only one set of these ratings. To ensure comparability with the official WMT results, we used the same set of ratings as the gold standard in this study.

A more rigorous evaluation would consider multiple rating sets from different raters and subject the gold standard itself to quality checks, including an assessment of inter-rater agreement. We did not use multiple rating sets or perform such checks in this paper. As a result, any noise or inconsistency in the selected raters' scores may have affected our evaluation of QE system performance, as it may also have affected the WMT23 rankings.

\section*{Acknowledgments}

We would like to thank Dr. Alon Lavie for insightful discussions and help with locating the WMT data. We also thank the contributors to the Conference on Machine Translation series (WMT) for making their data available to the public.

\bibliography{ROC_Garland_Berger}

\appendix

\section{ROC Analysis for Business Decisions}\label{sec:mainthresholds}

A common use case for QE systems is to triage translation segments for further review---for example, by human reviewers. ROC analysis enables users to select a QE score threshold that identifies the segments most likely to contain errors for review. Another common use case is selecting the highest-quality translations during data cleaning for machine learning. This use case is similar to the one described above, except that the objective is to identify high-quality translations rather than low-quality ones.

In both cases, the business decision ultimately comes down to balancing risk and cost. Below, we present several business decision scenarios using the triage use case as an example. The same solutions can be readily adapted to data cleaning and other QE-supported workflows. The methods described in this section can be implemented manually in Excel or automated using programming languages such as Python. Detailed guidance on code development is beyond the scope of this paper.

\subsection{QE-ROC Table}

In all of the scenarios below, a QE-ROC table is required. In practice, the QE-ROC table can be generated--explicitly or implicitly--using a spreadsheet program, a programming language, or statistical software with two minimal inputs: a ground-truth label and a QE score for each translation segment. Table~\ref{table:ROC} illustrates what a QE-ROC table may look like and how it is created.

\begin{table*}[bt!]
  \centering
  \begin{tabular}{|l|l|l|c|c|c|c|c|c|}
    \hline
    \textbf{Segment ID} & \textbf{Ground truth} & \textbf{QE score} & \textbf{TP} & \textbf{FN} & \textbf{FP} & \textbf{TN} & \textbf{TPR} & \textbf{FPR} \\
   \hline
    &  &  &  &  &  &  & \textbf{0.00} & \textbf{0.00} \\ 
\hline
5  & error    & 25  & 1 & 5 & 0 & 4 & 0.17 & 0.00 \\
\hline
9  & no error & 75  & 1 & 5 & 1 & 3 & 0.17 & 0.25 \\
\hline
6  & error    & 93  & 2 & 4 & 1 & 3 & 0.33 & 0.25 \\
\hline
1  & error    & 95  & 3 & 3 & 3 & 1 & 0.50 & 0.75 \\
\hline
2  & no error & 95  & 3 & 3 & 3 & 1 & 0.50 & 0.75 \\
\hline
4  & no error & 95  & 3 & 3 & 3 & 1 & 0.50 & 0.75 \\
\hline
8  & error    & 99  & 5 & 1 & 3 & 1 & 0.83 & 0.75 \\
\hline
10 & error    & 99  & 5 & 1 & 3 & 1 & 0.83 & 0.75 \\
\hline
3  & no error & 100 & 6 & 0 & 4 & 0 & 1.00 & 1.00 \\
\hline
7  & error    & 100 & 6 & 0 & 4 & 0 & 1.00 & 1.00 \\
\hline    &  &  &  &  &  &  & \textbf{1.00} & \textbf{1.00} \\
\hline  \end{tabular}
\caption{A sample QE-ROC table with 10 data points for illustration. In this example, the QE score is on a scale of 0-100, with 0 indicating the worst translation quality and 100 indicating perfect translation quality. The boldface values represent theoretical endpoints for the ROC curve and may or may not correspond to actual data points.}
  \label{table:ROC}
\end{table*}

First, the rows are sorted by the QE score -- from the worst score to the best score. Note that this is not necessarily from the highest to the lowest QE score, as shown in this example. Next, {\em TP, FN, FP}, and {\em TN} are calculated for each row using the QE score in that row as the threshold, specifically:
\begin{itemize}[nosep]
\item
{\em TP}: count of all rows that have "error" as the ground-truth label and QE scores worse than or equal to the QE score associated with the current row;
\item
{\em FN}: count of all rows that have "error" as the ground-truth label and QE scores better than the QE score associated with the current row;
\item
{\em FP}: count of all rows that have "no error" as the ground-truth label and QE scores worse than or equal to the QE score associated with the current row; and
\item
{\em TN}: count of all rows that have "no error" as the ground-truth label and QE scores better than the QE score associated with the current row.
\end{itemize}
Finally, {\em TPR} and {\em FPR} are calculated using Equations~(\ref{TPReq}) and (\ref{FPReq}) in Section~\ref{sec:definitions}, respectively.

A few notes about the QE-ROC table are warranted. 
First, a "best" or "better" QE score refers to a score indicating best or better translation quality. Depending on the QE system, this may be either a higher or lower score, as discussed in Section~\ref{sec:ROCsort}. Similarly, a "worst" or "worse" QE score refers to a score indicating worst or worse translation quality, which may also correspond to either a lower or higher score depending on the QE system.
Second, the QE system in this example outputs integer scores only, so some segments have identical QE scores; these segments are treated as ties, as explained in Section~\ref{sec:ROCsort}. Third, the ground-truth labels and QE scores shown in Table~\ref{table:ROC} are not necessarily the direct inputs used by ROC software libraries. They are presented in this format for ease of understanding. Finally, individual {\em FN} and {\em TN} counts are not required to construct the ROC curve and, for computational efficiency, need not be calculated separately. They are included in this illustrative table to support business decision-making.

\subsection{Assessing Risk When Resources Are Constrained}\label{sec:thresholds}

\begin{quote}
{\bf Scenario 1:} My organization only has the resources to have humans review {\em x}\% of all segments. How many error-containing segments will remain after human review, assuming that the human reviewers identify and correct all errors in the segments they review?
\end{quote}

{\bf Unpacking it:} We first need to identify which {\em x}\% of the segments will be sent for human review, and we can do this by finding a QE score threshold that separates the worst-scoring {\em x}\% segments from the rest. The segments sent for human review consist of {\em TP}s (error-containing segments identified as such by the QE system) and  {\em FP}s (error-free segments identified as error-containing by the QE system), and the segments not sent for human review consist of {\em TN}s (error-free segments identified as such by the QE system) and {\em FN}s (error-containing segments identified as error-free by the QE system). The human reviewers will correct the {\em TP}s and leave the {\em FP}s unchanged. The {\em TN}s and {\em FN}s will not be subject to human review and therefore remain unchanged. At the end of the process, the errors in the {\em FN}s will remain uncorrected, representing risk for the organization. Thus, to answer the question above, we need to estimate the number of {\em FN}s.

{\bf Step 1:} Create a QE-ROC table from a sample dataset that is representative of the data you are working on. 

{\bf Step 2:} Using the QE-ROC table, identify the QE score threshold that separates the {\em x}\% of segments that have the worst QE scores from the rest of the segments. 

{\bf Step 3:} From the QE-ROC table, read the {\em FN} count corresponding to the QE score threshold identified in Step 2.

{\bf Step 4:} Calculate the number of error-containing segments remaining per 100 total segments after human review:
\begin{displaymath}
\frac{FN \, \mbox{count from Step 3}}{\mbox{Total number of segments in sample}} \times 100
\end{displaymath}

By this point, we have obtained an estimate of how many error-containing segments remain per 100 segments, which can be used to assess residual risk. A confidence interval can also be estimated by first constructing a confidence band for the ROC curve, as described in Section~\ref{sec:bootstrap} and Appendix~\ref{app:bootstrap}, and then using the upper and lower limits of that band to derive the corresponding bounds for the percentage of error-containing segments remaining.

\subsection{Estimating Cost or Effort for Meeting a Risk Target}

\begin{quote}
{\bf Scenario 2:} My organization wants the final translation to have no more than {\em y} error-containing segments per 100 segments. What percentage of all segments do I need to send for human review, assuming that human reviewers identify and correct all errors in the segments they review?
\end{quote}

{\bf Unpacking it:} The remaining error-containing segments are the {\em FN}s---segments that contain errors but are classified by the QE system as error-free and thus never corrected because they are not reviewed by humans. We will need to select a QE score threshold at which the {\em FN} count is no more than y\%. Then, we can use that threshold to separate segments that will be sent for human review from the rest.

{\bf Step 1:} Create a QE-ROC table from a sample dataset that is representative of the data you are working on.

{\bf Step 2:} Calculate the maximum number of tolerable {\em FNs} in the sample:
\begin{displaymath}
y\% \cdot \mbox{Total number of segments in sample}
\end{displaymath}

{\bf Step 3:} From the QE-ROC table, read the QE score corresponding to the maximum number of tolerable {\em FN}s  calculated in Step 2. This is the QE score threshold that separates segments needing human review from the rest.

{\bf Step 4:} Count the number of segments with QE scores worse than or equal to the threshold identified in Step 3, and divide that number by the total number of segments in the sample to determine the percentage of all segments that need to be sent for human review.

A confidence interval for the result in this scenario can also be estimated using a method similar to that described at the end of the previous section.

\subsection{Setting the QE Score Threshold to Optimally Balance Risk and Cost}\label{sec:optimum}

The previous sections discussed how to meet either a cost target or a risk target. In practice, however, QE users may wish to satisfy both simultaneously. This raises an important limitation, because, as explained in Section~\ref{sec:addtlinfo}, any given QE system involves a trade-off between {\em FN}s and {\em FP}s. Although this balance can be adjusted by changing the QE score threshold, the achievable combinations of cost and risk are constrained by the QE system's performance limits. In some cases, a dual objective involving both cost and risk may simply be unattainable, particularly when the expectation is very demanding, such as achieving very low risk at very low cost.

Two natural questions therefore arise: What is the best performance that a given QE system can achieve for a particular use case, and how should the QE score threshold be set to achieve that optimal operating point? To answer these questions, three elements are needed: quantification of the {\em FN}-{\em FP} trade-off, the class ratio in the relevant data, and identification of the QE system's optimal operating point using the ROC curve and corresponding QE-ROC table.

To quantify the  {\em FN}-{\em FP} trade-off, we can assign a monetary value to each {\em FN}, and another to each {\em FP}. In the use case of triaging segments for further review, false negatives ({\em FN}s) represent risk because they are error-containing segments that are not reviewed and therefore remain uncorrected. False positives ({\em FP}s), by contrast, represent cost because these segments are in fact error-free and do not require correction, yet they are still sent for additional review. Similarly, in the use case of data cleaning, {\em FN}s represent risk because they introduce pollution into the resulting dataset, whereas {\em FP}s represent cost because they correspond to good data that are unnecessarily excluded. For example, if the monetary value associated with each {\em FN} is ten times that associated with each {\em FP}, then the trade-off is one {\em FN} to ten {\em FP}s, or 1:10.

Next, we determine the class ratio, that is, the ratio between the number of actual error-containing segments (the positive class) and the actual error-free segments (the negative class) in the translation. For example, if one-sixth of the segments in a machine translation output contain errors, then the {\em P:N} class ratio is 1:5, corresponding to one-sixth error-containing segments and five-sixths error-free segments.

Once both the {\em FN}-{\em FP} trade-off and the class ratio are known, the ROC curve can be used to identify the optimal operating point for the QE system. Specifically, we draw a line in ROC space with a slope that equals the quotient of the {\em FN}-{\em FP} trade-off and the {\em P:N}  class ratio. This line is known as the {\em iso-performance line}~\cite{FAWCETT2006861}. Then we shift this line, keeping it parallel, to a point on the ROC curve furthest toward the northwest of the ROC space. That point represents the operating condition that yields the optimal performance for the QE system under the specified {\em FN}-{\em FP}  trade-off and {\em P:N}  class ratio. The true positive rate ({\em TPR}) and false positive rate ({\em FPR}) at this point can then be read from the ROC curve, and the corresponding QE score threshold can be read from the QE-ROC table. This threshold is the one that optimizes performance of the QE system for the user's particular {\em FN}-{\em FP} trade-off scenario. 

\begin{figure}[htb!]
\centering
 \includegraphics[width=0.9\columnwidth]{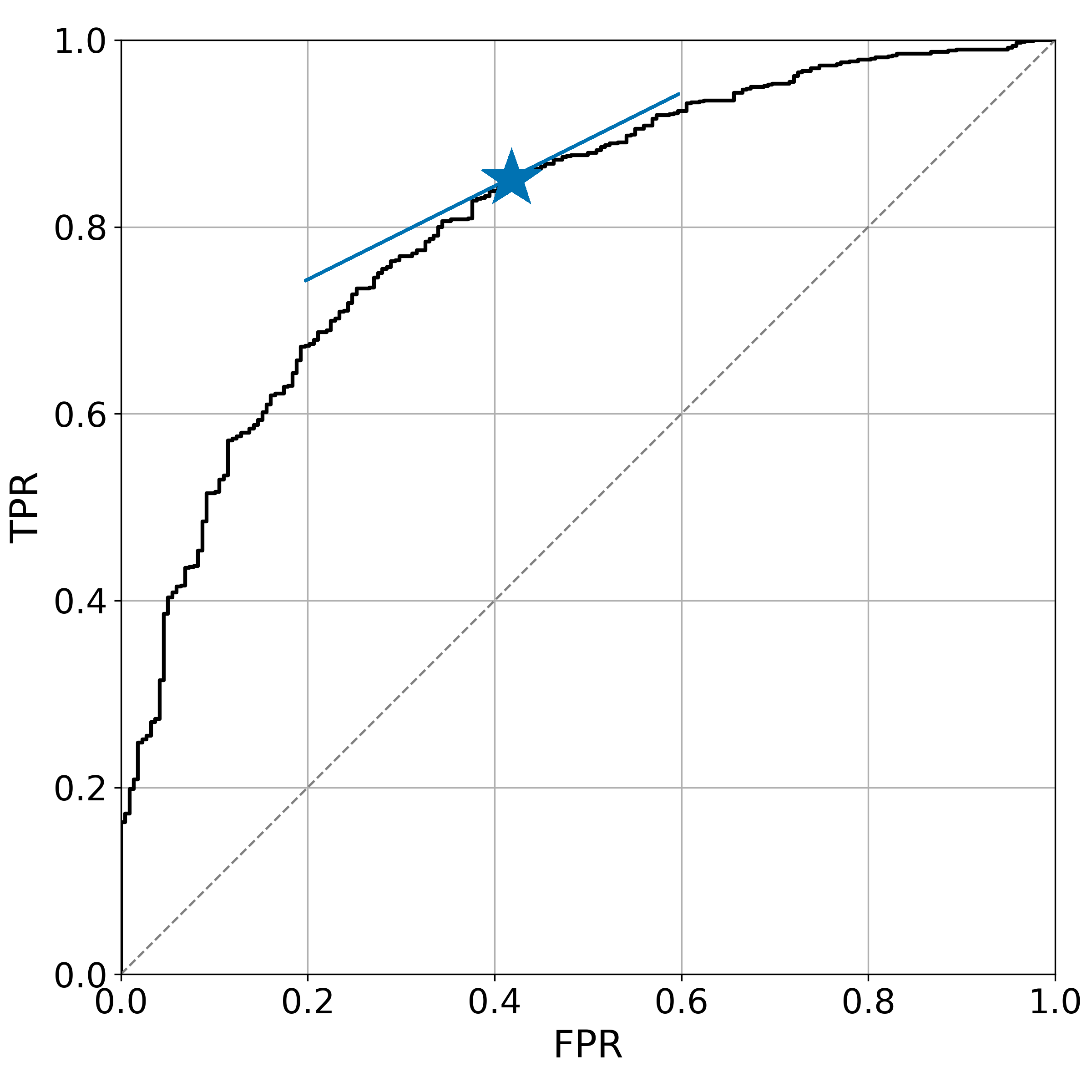} 
  \caption{The ROC curve of a QE system. The blue line has a slope of $\frac{1}{2}$ as calculated in the example and touches the ROC curve tangentially. The tangential point, marked by the blue star, indicates where the QE system achieves its best performance. }
  \label{fig:costriskbalance}
\end{figure}

In the example above, where the {\em FN}-{\em FP} trade-off is 1:10 and the {\em P:N}  class ratio is 1:5, as illustrated in Figure~\ref{fig:costriskbalance}, the slope of the iso-performance line is $\frac{\frac{1}{10}}{\frac{1}{5}} = \frac{1}{2}$. The blue star marks the point where the QE system can achieve its best performance under the specific conditions described in the examples.

A theoretical basis of this procedure is described in~\cite{Provost1998, Provost2001}.

This approach should be understood as an approximation, because {\em FN}s are not all equally consequential, and neither are {\em FP}s, for example, when risk should be weighted by error severity. Nevertheless, it provides a practical first-order estimate and is preferable to making threshold decisions without an explicit model of cost and risk.

It should also be noted that the foregoing analysis concerns only the trade-off between QE system classification errors. In practice, {\em FP}s are not the only costs. There are additional costs associated with the deployment of a QE system that need to be taken into consideration, including but not limited to: the cost of the system itself, such as subscription or licensing fees; the cost of learning, operating, maintaining, and monitoring the system; and the cost of training human users to apply it appropriately.

\subsection{Adaptation to Meet Different User Needs}

All solutions presented above can also be adapted to meet different user needs. For example, if an organization's experience suggests that human reviewers do not identify and correct all errors, the calculations can be adjusted accordingly, taking into account human inaccuracies.

Additionally, once the QE-ROC table has been created, it can be used to generate graphs and calculate values for other commonly used metrics, such as the precision-recall (PR) curve, PR-AUC (area under the precision-recall curve), and F1 score, using elements already included in the table. In other words, ROC analysis is not mutually exclusive with other evaluation methods; rather, it is complementary to them and shares much of the same data-processing workflow.

\section{Cautionary Notes---When Not to Use ROC Analysis}\label{sec:nouse}

\subsection{Ranking MT Systems}

It is tempting to use ROC curves or AUC values of one QE system to compare the performance of several MT systems. As already mentioned above in Section~\ref{sec:WMTrepro}, in particular when examining Tables~\ref{table:AUCzh} and~\ref{table:AUCde} for a given QE system, the area under the curve seems to decrease for better-performing MT systems. This is somewhat expected, because the errors that better-performing MT systems output are more subtle and therefore harder to detect. Nevertheless, ROC curves are neither able nor intended to pinpoint MT system performance. Instead, the output of one specific QE system should be utilized to evaluate the performance of different MT systems directly. 

\subsection{Non-Representative Samples}

In order to apply the ROC method to business decisions, as discussed in Appendix~\ref{sec:mainthresholds}, the studied dataset must be a representative sample of the type of data for which decisions are being made. This raises a related question: What is a representative sample, and how can one be obtained?

In the context of the business decision-making examples discussed in Appendix~\ref{sec:mainthresholds}, a representative sample is a dataset---typically a subset of the full dataset, or a separate but smaller dataset---that provides estimates of characteristics and results substantially similar to those that would be obtained from the full dataset.
In general, a representative sample should be sufficiently similar to the data it is intended to represent. This means that the sample and the working dataset should involve the same language pair, domain, and content type. In addition, the translations in both the sample and the working dataset should be produced by the same translator, whether human or machine. Furthermore, the sample size must be sufficiently large, as discussed below. If the sample dataset does not meet these conditions, the results of the ROC analysis may be misleading.

\subsection{Insufficient Sample Size}

In order to obtain meaningful ROC curves, not only the type of data, but also the size of the dataset is important. ROC curves are insensitive to the underlying class imbalance, that is, the ratio of the total number of positives $P$ to the total number of negatives $N$. In~\cite{Hanley1982MeaningAndUseROC}, under specific assumptions about the underlying distributions, it was shown that the precision of AUC depends on both $P$ and $N$, whereby the variance scales roughly with $\sim \frac{1}{P} + \frac{1}{N}$. 

In our case, the underlying distributions are unknown. Without making assumptions, only empirical analysis can determine whether a specific dataset is suitable for the ROC method. Our preliminary studies suggest that a minimum number of instances in each class is needed in order to obtain a reasonable ROC curve, even if the overall sample size is relatively large. This is in line with the aforementioned estimates in~\cite{Hanley1982MeaningAndUseROC}. The minimum number of instances per class seems to be roughly 50, but this number is of course highly dependent on the desired statistical significance.

Indicators of an unstable ROC curve are the confidence bands described in Section~\ref{sec:bootstrap} along with the confidence intervals for the corresponding AUC values. These can be obtained without making any assumptions about the distribution of the ground truth or the QE system classifier. In fact, even a near-random classifier, such as the QE system Random-sysname studied in Section~\ref{sec:WMTrepro}, can have relatively narrow confidence bands. However, if the resulting confidence bands are too wide, then we can say for certain that the ROC method should not be applied to the current dataset.

That is not to say that the ROC method should be abandoned completely in favor of other analyses. In the case of an insufficient sample size or a non-representative dataset, it is likely that these other metrics will not lead to statistically meaningful conclusions either. Instead, efforts should be made to augment the dataset with additional representative samples.

\section{Insensitivity to Changes in Class Ratio}\label{sec:imbalance}

Class ratio can be defined as the ratio of positive to negative instances, or $P:N$, and class imbalance occurs when one class substantially outnumbers the other. Many evaluation metrics are sensitive to changes in class ratio. As Fawcett~\cite{FAWCETT2006861} points out, any metric that draws on values from both columns of the confusion matrix is inherently affected by class ratio. Measures such as accuracy, precision, and F-score therefore vary as the class distribution changes, even when the underlying classifier performance remains unchanged. In contrast, ROC curves are insensitive to changes in class ratio. They use the complementary pairs in the confusion matrix---$TP/FN$ for the positive class and $FP/TN$ for the negative class---and are expressed proportionally to the number of positive and negative instances and are therefore independent of class ratio~\cite{Swets1988}.

The insensitivity of ROC curves to class imbalance holds only insofar as the underlying score distribution in each class remains stable~\cite{RICHARDSON2024100994}. In the QE setting, this means that the distribution of translation error types and severities, as well as the distribution of characteristics of error-free segments, should remain broadly similar across the datasets being compared.

The relative insensitivity of ROC analysis to changes in class ratio is an attractive property and has been widely recognized in many fields. However, how fully this property applies to translation quality evaluation, and to QE in particular, requires further investigation. The reason is that, unlike in many other application domains, neither the positive nor negative class in translation QE is homogeneous.

In other fields, such as defect detection in manufacturing or disease screening in epidemiology, the target condition is often relatively well defined and internally consistent. By contrast, in translation quality evaluation, the categories of "errors" or "non-errors" each encompass a heterogeneous set of instances that exhibit a wide variety of characteristics. Translation errors may differ substantially in type, severity, and linguistic manifestation. Some error types may be readily captured by a QE system, whereas others may be much more difficult to detect. Likewise, error-free translations may also vary considerably in terms of length, structure, and other features, and may also differ in how readily they can be identified as such by a QE system. At present, research remains limited on which types of errors and error-free translations QE systems can detect reliably.

Such heterogeneity and variation matter because they limit the extent to which ROC's insensitivity to class imbalance can be leveraged. In translation quality estimation, it cannot generally be assumed that datasets with different class ratios share the same underlying error distribution, nor can it be assumed that the distribution of characteristics of error-free segments remains constant. For example, if two MT systems produce translations with different class ratios, i.e., one MT system produces more errors than another, the difference is unlikely to be purely quantitative; the two systems may produce different types of errors, different severities of errors, or different mixtures of error types and severities. The same concern applies across language pairs, domains, and content types. Similarly, when the proportion of correctly translated segments changes across datasets, the distribution of characteristics within the error-free class may also shift.

Further exploration of how ROC's insensitivity to changes in class ratio can be applied in translation QE would require  investigation of the distribution of the underlying data in each class, which goes beyond the scope of the present paper.

\section{Nonparametric Stratified Bootstrap Resampling}~\label{app:bootstrap}

Below we briefly summarize the bootstrap resampling algorithm we used to obtain confidence bands and intervals. The mathematical details of this bootstrap resampling technique are discussed, for example, in~\cite{Rboot,HorvathHorvathZhou2008,WuMartinKacker2016}.

\begin{algorithm}[h]
\caption{Nonparametric Stratified Bootstrap Resampling for ROC Confidence Bands}\label{alg:boot}
\DontPrintSemicolon
Begin with empty set of resampled ROC curves\;
Begin with empty list of resampled AUC values\;
Compute {\em FPR} grid value granularity $\gets ~\sim \frac{1}{N}$\;
\For{$b\gets 1$ \KwTo $B$}{
    Resample positive segments $P$ times with replacement\;
    Resample negative segments $N$ times with replacement\;
    Combine both samples into a full sample set with the same class imbalance and total number of segments as the original ROC curve\;
    Compute ROC curve and AUC value\;
    Append AUC to list of AUC values\;
    Interpolate ROC curve onto the {\em FPR} grid\;
    Add interpolated ROC curve to set of resampled ROC curves\;
}
Compute pointwise confidence bands for each {\em FPR} value on the grid\;
Compute confidence intervals from $B$ AUC values in list of AUC values\;
\end{algorithm}

As shown in Algorithm~\ref{alg:boot}, we first separate the segments as classified per the ground truth, into positive segments, in total $P$, and negative segments, in total $N$, along with their classifier scores. We then iterate the following $B$ number of times: We sample the positive segments $P$ times with replacement, we sample the negative segments $N$ times with replacement, and then combine the two samples. This gives us a stratified sample set with the same class imbalance and total number of segments as for our original ROC curve. For this resampled dataset, we compute the ROC curve as before. After iterating this procedure $B$ times, we have a set of $B$ resampled ROC curves from which we can compute confidence bands. However, since the two-dimensional points ({\em FPR,TPR}) do not necessarily have the same {\em FPR} values, we have to first interpolate each of the $B$ resampled curves onto the same {\em FPR} grid. The granularity of this grid is chosen to be of the order of $\sim \frac{1}{N}$. After interpolation, we have $B$ resampled interpolated ROC curves with the same {\em FPR} values on the x-axis. From these curves, we can now finally compute our confidence bands pointwise for each {\em FPR} value. For example, to get a 95\% confidence band, we would compute the 2.5\textsuperscript{th} percentile for the lower band and the 97.5\textsuperscript{th} percentile for the upper band.  

To compute the confidence intervals for the AUC values, we follow a similar procedure, also shown in Algorithm~\ref{alg:boot}. We resample $B$ times, as explained above. From these $B$ resampled ROC curves, we compute $B$ AUC values directly, before the confidence band interpolation step above. From this set of $B$ resampled AUC values, we can compute the standard confidence intervals in the usual way. 

\end{document}